\documentclass[runningheads]{llncs}
\usepackage[T1]{fontenc}
\usepackage{multirow}
\usepackage{graphicx}
\usepackage[pagebackref=true,breaklinks=true,colorlinks,bookmarks=false]{hyperref}
\usepackage{booktabs}
\usepackage{tabularx}
\usepackage{placeins}
\usepackage{array}
\usepackage{marvosym}
\usepackage{tcolorbox}

\begin{document}
\title{Multimodal Medical Image Parsing with Task-Specific Low-Rank Adapters}
\titlerunning{Medical Image Parsing with Task-Specific Adapters}
\author{Taha Sabir\inst{1}\orcidID{0009-0002-4870-0017} \and
Alberto Montero Molina\inst{2,\dagger}\orcidID{0009-0008-6798-0943} \and
Thinh Quang Ly\inst{3,\dagger}\orcidID{0009-0006-9039-5887} \and
Veronika Dmitrenko\inst{4,\dagger}\orcidID{0009-0006-4636-7440} \and
Mohammad Yaqub\inst{5}\orcidID{0000-0001-6896-1105} \and
Tausifa Jan Saleem\inst{5}\textsuperscript{(\Letter)}\orcidID{0000-0002-0827-0043}}
\authorrunning{T. Sabir et al.}
\institute{Department of Computer Science, Gettysburg College, Gettysburg, PA 17325, USA \and
Higher School of Computing (ESCOM), Instituto Polit\'ecnico Nacional (IPN),
Mexico City 07738, Mexico \and
Department of Computer Science and Engineering, Michigan State University,
East Lansing, MI 48824, USA \and
Faculty of Computer Science and Faculty of Economic Sciences, HSE University, Moscow, Russia \and
Computing and Mathematical Sciences Division, Mohamed bin Zayed University of Artificial Intelligence, Abu Dhabi, United Arab Emirates\\
\email{tausifa.saleem@mbzuai.ac.ae}\\
\textsuperscript{$\dagger$}These authors contributed equally to this work.}
\maketitle              
\begin{abstract}
Adapting a single vision-language model to medical image parsing requires training choices that reflect the demands of each task. We study task-specific adaptation with Qwen3.5-9B for FLARE 2026 Task 3, covering six task families and seven imaging modalities. Our system couples a shared, frozen 4-bit backbone with LoRA adapters selected from the input question. We customize adapter ranks and prompts, rebalance rare classification targets, and supplement regression training with dermatology classification examples. On the fixed internal holdout, the task-specific adapters achieve a dataset-macro balanced accuracy of 0.7791 for classification, a dataset-macro micro-F1 of 0.6286 for multi-label classification, a counting mean absolute error (MAE) of 11.73, a dataset-macro detection F1 of 0.8914, a dataset-macro regression MAE of 9.78, and a report-generation CRIMSON score of 0.8204. On the organizer's 1,703-example hidden-validation set, the submitted system achieves 0.8496 classification balanced accuracy, 0.5338 sample-averaged multi-label F1, and 14.5950 regression MAE. The final 30.1 GB container averages 3.1 seconds per case, including model loading, and produces identical outputs across repeated verification runs. The benefits vary across tasks, supporting the use of task-specific adapter configurations and image budgets within a shared medical vision-language model. Code is available at \url{https://github.com/TahaSabir0/FLARE26-Hexapus}.

\end{abstract}

\section{Introduction}

Unlike conventional visual question answering (VQA), where the expected response is primarily textual, medical image parsing requires a general medical vision-language model (VLM) to interpret heterogeneous visual inputs and produce task-appropriate outputs ranging from categorical and multi-label predictions to numerical estimates, bounding boxes, and free-form reports, across modalities with different visual characteristics and anatomical morphologies. Medical VLMs range from biomedical dialogue systems such as LLaVA-Med~\cite{li2023llavamed} to models pretrained across clinical imaging domains such as MedGemma~\cite{sellergren2025medgemma}, and benchmarks compare generalist and specialist performance across clinical domains~\cite{zhong2025generalist}. Med-MoE~\cite{jiang2024medmoe} illustrates the need to combine discriminative and generative medical tasks, while a systematic task-transfer study~\cite{sachdeva2026tasktransfer} shows that fine-tuning one visual capability can unpredictably improve or degrade others.

FLARE 2026 Task 3~\cite{flare2026task3} provides a practical setting for studying this problem. After data cleaning, its 51,154 question-answer pairs and 50,996 images span 19 datasets, eight imaging modalities, and seven task types; excluding unscored instance detection and endoscopy, our evaluation covers 17 datasets, seven modalities, and six scored task families: classification, multi-label classification, counting, detection, regression, and report generation. A single shared adaptation may suffer from interference between these tasks, whereas a separate fine-tuned VLM per task would substantially increase training, storage, and deployment costs.

\noindent\textbf{Related Work.}
General-domain VLMs such as BLIP-2~\cite{li2023blip2} and Qwen2.5-VL~\cite{bai2025qwen25vl} align visual and language representations through lightweight bridging modules or dynamic-resolution processing, while medical systems including LLaVA-Med~\cite{li2023llavamed}, HuatuoGPT-Vision~\cite{chen2024huatuogptvision}, and MedGemma~\cite{sellergren2025medgemma} add medical image--text training or domain-specific visual representations. Yet parsing outputs differ in visual evidence, supervision, and decoding constraints. Previous challenge solutions to FLARE 2025 Task 5~\cite{ma2025flare} use different forms of specialization: QwenVLConnector~\cite{nguyen2025qwenvlconnector} fuses multi-level visual features, ME-VLIP~\cite{shaaban2025mevlip} routes questions among task-specific QLoRA adapters and a multi-task fallback, and Sun et al.~\cite{sun2025multitask} use a medically pretrained vision encoder with a shared LoRA-adapted language model. These approaches show that broad task coverage is possible but leave open how task-specific adaptation, prompting, and visual resolution should be allocated within one shared, deployable backbone. LoRA~\cite{hu2022lora}, QLoRA~\cite{dettmers2023qlora}, and LLaMA-Factory~\cite{zheng2024llamafactory} make such modular specialization practical.

\noindent\textbf{Contributions.}
Our system uses Qwen3.5-9B~\cite{qwen2026qwen35} with a question-only GLiClass router~\cite{stepanov2025gliclass} (DeBERTa-V3~\cite{he2023debertav3}) selecting one of six specialist adapters or a general fallback on one shared 4-bit backbone. Our contributions are:
\begin{itemize}
    \item We develop and validate a parameter-efficient system for six medical image-parsing tasks using one quantized Qwen3.5-9B backbone and lightweight task-specific LoRA adapters.
    \item We compare the adapted system with its zero-shot Qwen3.5-9B backbone on an internal holdout and public validation data.
    \item We compare specialist and general adapters, measure task-dependent resolution effects, and analyze errors hidden by aggregate metrics.
\end{itemize}

\section{Method}
\label{sec:method}

\begin{figure}[htbp]
\centering
\includegraphics[width=1\textwidth]{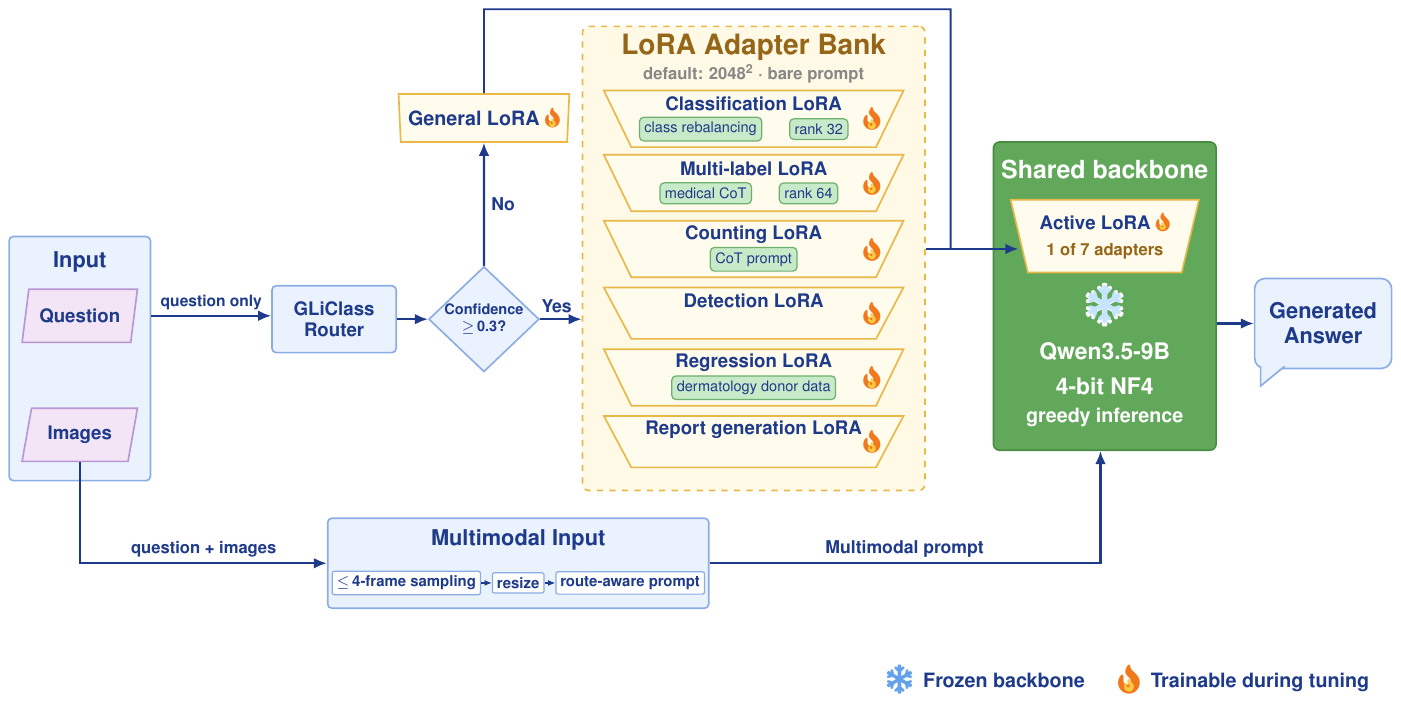}
\caption{Routed inference with a shared quantized backbone. The question selects a specialist or the general fallback, together with the image budget and prompt used for answer generation.}
\label{fig:pipeline}
\end{figure}

We use one quantized Qwen3.5-9B backbone with task-specific LoRA adapters. A question-only router selects the adapter, image budget, and prompt for each case (Fig.~\ref{fig:pipeline}).

\subsection{Data preparation and image handling}
\label{sec:data_preparation}

We audited the official FLARE 2026 Task 3 2D data~\cite{flare2026task3} and excluded 73 records that failed task-specific validity checks: 47 endoscopy records with invalid labels, 17 BUSI detection records with inverted or zero-area boxes, seven periapical records with contradictory labels, one duplicated periapical record, and one Boneresorption record with an extreme target value of 2,085 (dataset median: 28). We also excluded all 1,248 instance-detection samples because this task was not scored, and omitted endoscopy from adapter training because of its few usable examples, long multi-frame inputs, and noisy labels.

Following ME-VLIP~\cite{shaaban2025mevlip}, we expanded each multi-image training example into separate single-image examples, repeating the question and answer for each image. Questions sharing an image remained in the same partition. In earlier Qwen3-VL-8B experiments, grouping three neojaundice images reduced balanced accuracy in two complementary train--validation folds; this motivated, but does not validate, the policy for Qwen3.5-9B. At inference, the system accepts up to four images and selects four evenly spaced images from longer sequences.

Image preprocessing consists of RGB conversion and aspect-ratio-preserving resizing when an image exceeds its route's pixel-area cap. The cap is a maximum area, not a forced square shape. We used no additional image transformations, unlabeled examples, or partial labels.

\subsection{Shared backbone and task routing}

All routes use one Qwen3.5-9B backbone loaded with 4-bit NF4 quantization and bfloat16 computation. The pretrained backbone remains frozen. Six task-specific LoRA adapters and one general adapter, trained across all scored tasks, are loaded into the same model; switching tasks changes the active adapter without reloading the backbone.

We reuse the GLiClass classifier released with ME-VLIP~\cite{shaaban2025mevlip}, without further training. It uses a DeBERTa-v3-small encoder to predict a task label and confidence from the question text alone. The corresponding specialist is activated when the top confidence is at least 0.3 and the label has a configured route; otherwise, the general adapter is used. Supplied \texttt{TaskType} and \texttt{Modality} metadata do not influence routing.

\subsection{Task-specific adaptation}

We train each adapter by supervised fine-tuning, optimizing LoRA parameters with loss applied only to the answer tokens. Table~\ref{tab:route_config} lists the deployed configurations. In the table, the full labeled pool comprises $A\cup B$, the internal holdout, and \texttt{validation-public}; prompt variants are defined in Section~\ref{sec:prompting}. Shared optimization settings are given in Section~\ref{sec:training_protocols}.

\begin{table}[!htbp]
\centering
\caption{Per-route configuration of the proposed system. Pixel values denote maximum image-area budgets, not forced square resizing.}
\label{tab:route_config}
\scriptsize
\setlength{\tabcolsep}{3pt}
\renewcommand{\arraystretch}{1.18}
\begin{tabularx}{\textwidth}{@{}p{2.35cm}X p{1.05cm} p{1.15cm} p{1.45cm} p{1.75cm}@{}}
\toprule
Route & Data scope & Pixels & LoRA r/$\alpha$ & Prompt & Output \\
\midrule
Classification
& Full labeled pool, oversampled
& $2048^2$
& 32/64
& Bare
& Label string \\

Multi-label classification
& Full labeled pool
& $2048^2$
& 64/128
& Medical
& Label set \\

Counting
& Full labeled pool
& $2048^2$
& 8/16
& Counting
& Numeric string \\

Detection
& Full labeled pool
& $2048^2$
& 8/16
& Bare
& Nested boxes \\

Regression
& A$\cup$B + \texttt{bcn20000} donor mixture
& $2048^2$
& 8/16
& Bare
& Numeric string \\

Report generation
& A$\cup$B
& $768^2$
& 8/16
& Bare
& Free text \\

General adapter
& A$\cup$B, all scored tasks
& $768^2$
& 8/16
& Bare
& Task-dependent \\
\bottomrule
\end{tabularx}
\end{table}

\FloatBarrier

Preliminary Qwen3-VL-8B experiments informed the LoRA~\cite{hu2022lora} ranks retained for Qwen3.5-9B: 32 for classification, 64 for multi-label classification, and 8 for the remaining adapters, with $\alpha=2r$ throughout.

For classification, we use capped oversampling of rare dataset--answer groups. A group containing $n<250$ training examples is expanded by duplication to $\min(6n,250)$ examples; larger groups are unchanged. For regression, we combine 5,265 examples from $A\cup B$ with 5,000 \texttt{bcn20000} dermatology-classification examples. The $A/B$ partitions and internal holdout are defined in Section~\ref{sec:datasets}.

Adapters trained on $A\cup B$ provide the internal comparisons. For deployment, classification, multi-label classification, counting, and detection were subsequently refitted on the full labeled development pool. Regression, report generation, and the general fallback retain their $A\cup B$ adapters.

\subsection{Prompting and inference}
\label{sec:prompting}

Each user message contains indexed image markers followed by the original question. Counting and multi-label classification also use the instructions below, placed before the question. They are inspired by chain-of-thought prompting~\cite{wei2022cot}, but training supervises only final answers. The remaining routes use the bare image--question prompt.

\begin{tcolorbox}[title={Task-specific prompts},colback=black!3,colframe=black!70,
  colbacktitle=black!70,coltitle=white,fonttitle=\bfseries,
  boxrule=0.6pt,arc=2pt,left=8pt,right=8pt,top=7pt,bottom=7pt]
\small\raggedright
\textbf{Counting}\par\smallskip
\textbf{System}\par
You are a medical image analysis expert. Think carefully before answering.\par\smallskip
\textbf{Task instruction}\par
Methodically count each object in the image (scan left-to-right, row by row). Write the final count as a single integer.
\tcblower
\textbf{Multi-label classification}\par\smallskip
\textbf{System}\par
You are an expert radiologist and medical image analyst specialising in X-ray, CT, MRI, ultrasound, pathology, endoscopy, and microscopy. Think carefully and systematically before answering.\par\smallskip
\textbf{Task instruction}\par
For each listed finding, assess whether it is clearly visible in the image. List only confirmed findings separated by commas.
\end{tcolorbox}

Generation uses greedy decoding (\texttt{do\_sample=False}), at most 512 new tokens, fixed end-of-sequence and padding identifiers, and Qwen3.5 thinking mode disabled. Post-processing removes any thinking tags, retains the final line for classification, normalizes multi-label delimiters, extracts detection boxes and numeric answers, and retains report text. If no numeric value is found, the numeric parser returns zero. We apply no calibration, re-querying, or manual answer correction.

The container packages the backbone, adapters, tokenizer, and router for fully offline execution. Its runtime, size, and repeated-run checks are reported in Section~\ref{sec:deployment}.

\section{Experiments}
\subsection{Dataset and evaluation measures}
\label{sec:datasets}

\noindent\textbf{Dataset.}
The cleaned source pool for FLARE 2026 Task 3~\cite{flare2026task3} contained 51,154 question--answer pairs from the organizer training and public-validation data. After the exclusions described in Section~\ref{sec:data_preparation}, our evaluation covers 17 datasets, seven imaging modalities, and six scored task families.

\begin{figure}[!htbp]
\centering
\includegraphics[width=0.95\textwidth]{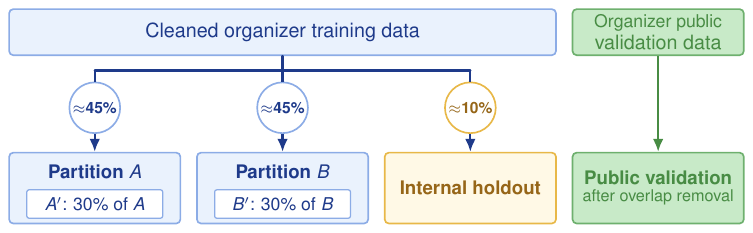}
\caption{Image-grouped development partitions and configuration-screening subsets. Percentages are approximate; $A'$ and $B'$ each contain 30\% of their parent partition. Public-validation images overlapping the organizer training data were removed.}
\label{fig:data_splits}
\end{figure}

Following recommended cross-validation practice for medical-imaging AI~\cite{bradshaw2023crossvalidation}, we partitioned the organizer training data by image group, stratified by dataset and task combination, into an approximately 10\% internal holdout and two 45\% subsets, $A$ and $B$ (Fig.~\ref{fig:data_splits}). Smaller subsets $A'\subset A$ and $B'\subset B$, each containing 30\% of its parent partition, supported inexpensive preliminary screening before selected configurations were evaluated on the full $A/B$ partitions. Full $A\rightarrow B$ and $B\rightarrow A$ experiments were run where available. These partitions were fixed before expanding multi-image examples into single-image training rows.

Adapted models in the internal and public-validation comparisons are trained on $A\cup B$; public-validation images overlapping the organizer training data were excluded. Although internal-holdout images were excluded from these adapters' training, holdout results informed some development choices. We therefore treat the holdout results as development comparisons rather than independent estimates of generalization~\cite{varma2006bias}. Selected deployment routes were later refitted with the holdout and public-validation data (Table~\ref{tab:route_config}); scores on these previously seen cases are not reported as held-out performance.

\FloatBarrier

\noindent\textbf{Evaluation.}
We use balanced accuracy (mean class recall) for classification and micro-averaged F1 for multi-label classification. Detection uses F1 with box matching at intersection-over-union $>0.5$. Counting and regression use mean absolute error (MAE), and report generation uses the language-model-based CRIMSON metric~\cite{baharoon2026crimson}.

We evaluate the internal and public splits separately, computing each metric per dataset and reporting its unweighted mean within each task. Counting and report generation each contain one evaluated dataset. Excluding endoscopy, internal/public classification covers ten/six datasets, multi-label classification two/one, and detection three/two. Regression has two internal datasets and no public cases. Organizer hidden-validation scores (Section~\ref{sec:hidden_val}) instead pool cases within each task. Differences in dataset composition and aggregation prevent direct comparison with our dataset-macro results.

\noindent\textbf{Statistical uncertainty.}
Paired 95\% confidence intervals in the configuration ablations use 10,000 bootstrap resamples of matched question--answer cases, sampled with replacement within each dataset using the same sampled indices for both configurations. For the general-versus-specialist regression comparison, interval bounds are the 2.5th and 97.5th percentiles of the resampled MAE differences. This case-level resampling treats question--answer cases as independent; the intervals do not account for dependence between questions sharing an image or variation across training seeds.

\noindent\textbf{Comparison protocols.}
\label{sec:baseline_protocol}
We compare $A\cup B$ adapters with two zero-shot baselines on the same splits. The first uses the deployed prompts, answer prefixes, extraction rules, and 512-token limit. The format-assisted baseline adds explicit output-format instructions, removes the answer prefix, changes answer extraction, and allows 1,024 tokens. Both retain the same 4-bit Qwen3.5-9B backbone, greedy decoding, disabled thinking mode, and task-specific image budgets ($768^2$ for reports, $2048^2$ otherwise). Thus, their difference measures a combined inference-protocol change, not a prompt-only effect. These evaluations select configurations using known task assignments and do not measure router accuracy or end-to-end container performance. Report generation uses only the first baseline; comparisons retain reports scored successfully for both models (780 internal and 1,943 public cases).

\subsection{Implementation details}
\subsubsection{Environment settings}

Table~\ref{table:env} summarizes the single-GPU development environments used with LLaMA-Factory.

\begin{table}[!htbp]
\caption{Development environment.}
\label{table:env}
\centering
\small
\begin{tabular}{ll}
\hline
\textbf{Component} & \textbf{Setting} \\
\hline
System & Ubuntu 22.04.5 LTS (A5000 workstation) \\
CPU / RAM & 32 cores / 256 GB (A5000 workstation) \\
GPUs & NVIDIA RTX A5000 (24 GB) \\
     & NVIDIA RTX PRO 6000 Blackwell (96 GB) \\
Framework & PyTorch 2.11.0, CUDA 12.8 \\
Training library & LLaMA-Factory 0.9.5 \\
Dependencies & Transformers 5.16.1, PEFT 0.18.1 \\
             & bitsandbytes 0.50.2 \\
Code & \url{https://github.com/TahaSabir0/FLARE26-Hexapus} \\
\hline
\end{tabular}
\end{table}

\subsubsection{Training protocols}
\label{sec:training_protocols}

We applied QLoRA to all linear layers of the language model while keeping the pretrained weights, vision encoder, and multimodal projector frozen. Supervised fine-tuning used next-token cross-entropy loss on answer tokens only. Table~\ref{table:training} lists the shared settings. Runs used one GPU and an effective batch size of 8: batch size 2 with four accumulation steps on Blackwell, or batch size 1 with eight steps on the A5000.

An earlier single-seed Qwen3-VL-8B classification screen favored three epochs over two or four (0.5964 versus 0.5697 and 0.5876 dataset-macro balanced accuracy on $A'\rightarrow B'$ at $768^2$). We transferred this schedule to Qwen3.5-9B without treating it as optimal for the final backbone or every task.

\begin{table}[!htbp]
\caption{Shared training settings for the QLoRA adapters.}
\label{table:training}
\centering
\small
\begin{tabular}{ll}
\hline
\textbf{Parameter} & \textbf{Setting} \\
\hline
Epochs & 3 \\
Effective batch size & 8 \\
Optimizer & AdamW \\
Initial learning rate & $2\times10^{-4}$ \\
Learning rate schedule & Linear decay, 3\% warm-up \\
Maximum sequence length & 8,192 tokens \\
Compute precision & BF16 \\
Gradient checkpointing & Enabled \\
Random seed & 1337 \\
\hline
\end{tabular}
\end{table}

Each adapter was evaluated at its training resolution using the final checkpoint, without early stopping, extra augmentation, or patch sampling. All reported Qwen3.5-9B experiments use one random seed; Section~\ref{sec:datasets} describes configuration selection.

\section{Results and discussion}
\label{sec:results}

\subsection{Results on the validation sets}

\noindent\textbf{Internal and public validation.}

\begin{figure}[!t]
\centering
\includegraphics[width=\textwidth]{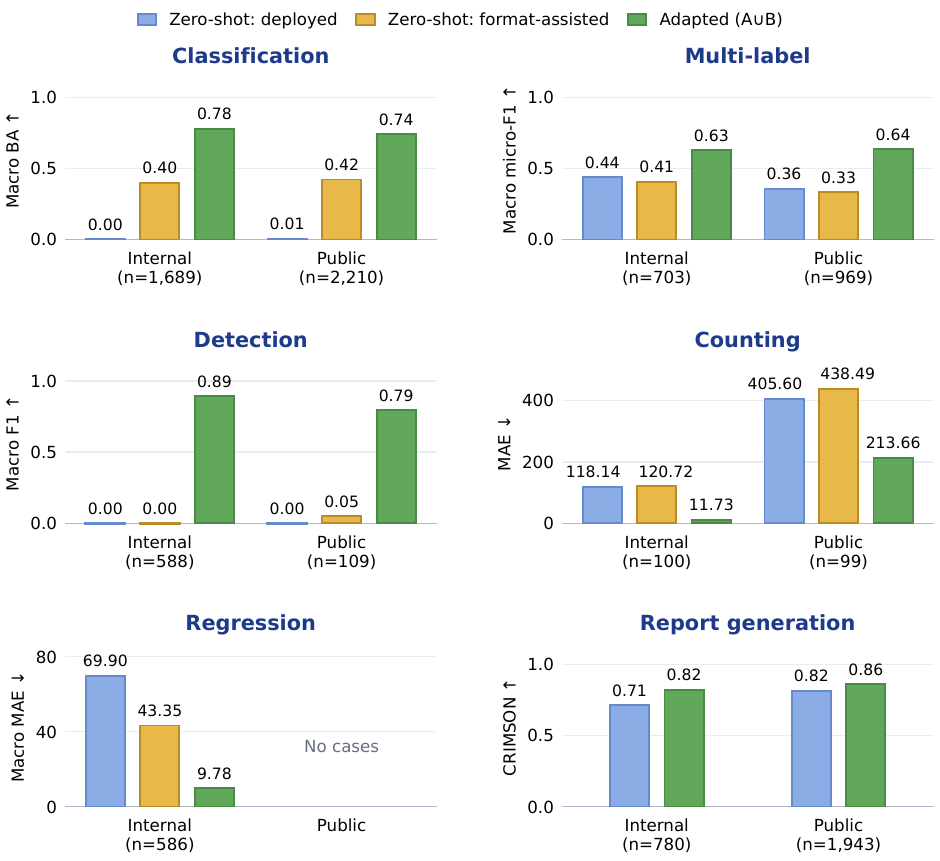}
\caption{Zero-shot and adapted Qwen3.5-9B on internal and public-validation
data, using the protocols in Section~\ref{sec:datasets}. BA denotes balanced
accuracy; multi-label F1 is micro-averaged within each dataset. Task scores
are dataset macros, except single-dataset counting and report generation.
Report means use the same successfully scored cases for both models; no
format-assisted report run was performed. Public validation has no regression
cases. Arrows indicate the better direction.}
\label{fig:baseline_comparison}
\end{figure}

Figure~\ref{fig:baseline_comparison} compares the route-matched $A\cup B$
adapters with the two zero-shot protocols. These adapters exclude internal
holdout and public-validation images from training; they are proxies for
deployment routes subsequently refitted on the full labeled pool.
Table~\ref{tab:final_dataset_results} gives the internal per-dataset results.

Output conventions materially affect zero-shot performance. Format assistance
raises classification balanced accuracy from 0.0035 to 0.3982 internally and
from 0.0055 to 0.4229 publicly; adaptation then reaches 0.7791 and 0.7418.
The same protocol change does not help multi-label classification or counting,
and near-zero detection scores remain sensitive to coordinate conventions and
answer extraction rather than measuring localization alone.

Adaptation improves every evaluated task over the available zero-shot protocols,
although large errors remain. Relative to the deployed-settings baseline,
counting MAE falls from 118.14 to 11.73 internally and from 405.60 to 213.66
publicly, where dense cell images remain difficult. On jointly scored reports,
CRIMSON rises from 0.7141 to 0.8208 internally and from 0.8151 to 0.8622 publicly.

\begin{table}[!htb]
\centering
\caption{Per-dataset results of the route-matched Qwen3.5-9B adapters on the
internal holdout. The report score uses all 781 adapted-model scores;
Fig.~\ref{fig:baseline_comparison} uses the 780 reports jointly scored with
zero-shot.}
\label{tab:final_dataset_results}
\scriptsize
\setlength{\tabcolsep}{3pt}
\renewcommand{\arraystretch}{1.00}
\begin{tabular}{@{}p{2.6cm}p{1.8cm}lrrr@{}}
\hline
Task & Dataset & Metric & $n$ & Resolution & Qwen3.5-9B \\
\hline
Classification & CMMD & BA & 357 & $2048^2$ & 0.5970 \\
 & BCN20000 & BA & 286 & $2048^2$ & 0.6900 \\
 & Dental & BA & 420 & $2048^2$ & 0.6700 \\
 & Neojaundice & BA & 60 & $2048^2$ & 0.8100 \\
 & Bone marrow & BA & 72 & $2048^2$ & 0.8660 \\
 & BUS-UCLM & BA & 52 & $2048^2$ & 0.9190 \\
 & Fundus & BA & 138 & $2048^2$ & 0.6870 \\
 & IUGC & BA & 131 & $2048^2$ & 0.9300 \\
 & Retino & BA & 111 & $2048^2$ & 0.7610 \\
 & BUSI & BA & 62 & $2048^2$ & 0.8610 \\
\hline
Multi-label & ChestDR & F1 & 388 & $2048^2$ & 0.5625 \\
 & Periapical & F1 & 315 & $2048^2$ & 0.6947 \\
\hline
Counting & NeurIPS22Cell & MAE & 100 & $2048^2$ & 11.7300 \\
\hline
Detection & IUGC & F1 & 516 & $2048^2$ & 0.9981 \\
 & BUSI-det & F1 & 50 & $2048^2$ & 0.8317 \\
 & BUS-UCLM-det & F1 & 22 & $2048^2$ & 0.8444 \\
\hline
Regression & Boneresorption & MAE & 70 & $2048^2$ & 17.1600 \\
 & IUGC & MAE & 516 & $2048^2$ & 2.4100 \\
\hline
Report generation & IU-XRay & CRIMSON & 781 & $768^2$ & 0.8204 \\
\hline
\end{tabular}
\end{table}

\noindent\textbf{Organizer hidden validation.}
\label{sec:hidden_val}
The hidden-validation submission was scored on 1,703 cases spanning
classification, multi-label classification, and regression
(Table~\ref{tab:hidden_val}). The official overall score was 0.4825. These
metrics are computed over each task's full scoring surface; in particular, the
challenge's sample-averaged multi-label F1 is distinct from the per-dataset micro-F1 used in our internal
and public-validation comparisons.

\begin{table}[!htbp]
\centering
\caption{Official hidden-validation performance of the submitted Qwen3.5-9B
system. Scores are pooled by task.}
\label{tab:hidden_val}
\scriptsize
\setlength{\tabcolsep}{3pt}
\renewcommand{\arraystretch}{1.05}
\begin{tabular}{@{}llrr@{}}
\hline
Task & Metric & $n$ & Score \\
\hline
Classification & Balanced accuracy $\uparrow$ & 1,126 & 0.8496 \\
Multi-label classification & Sample-averaged F1 $\uparrow$ & 477 & 0.5338 \\
Regression & MAE $\downarrow$ & 100 & 14.5950 \\
\hline
\end{tabular}
\end{table}

\subsection{Configuration ablations}
\subsubsection{General versus task-specific adapters.}

Task-specific configurations produced uneven benefits over the shared general
adapter (Table~\ref{tab:configuration_ablations}(a)). All rows compare matched
$768^2$ input. Specialists score higher on four of six tasks, including the
largest scored slice, classification, while the multi-label and
report-generation comparisons show no clear difference.

\begin{table}[!htb]
\centering
\caption{Configuration ablations on the internal holdout. (a) General adapter
versus task specialists at matched $768^2$ input. (b) Resolution comparisons
for task specialists. Regression is the mean MAE over two datasets.}
\label{tab:configuration_ablations}
\scriptsize
\setlength{\tabcolsep}{3pt}
\renewcommand{\arraystretch}{1.05}
\begin{tabular}{@{}llrr@{}}
\hline
\multicolumn{4}{@{}l}{\textbf{(a) General versus task-specific adapters}} \\
Task & Metric & General & Specialist \\
\hline
Classification & BA $\uparrow$ & 0.7348 & \textbf{0.7695} \\
Multi-label classification & F1 $\uparrow$ & \textbf{0.6157} & 0.6130 \\
Counting & MAE $\downarrow$ & 14.3400 & \textbf{13.4500} \\
Detection & F1 $\uparrow$ & 0.8876 & \textbf{0.9194} \\
Regression & MAE $\downarrow$ & 12.1600 & \textbf{10.1800} \\
Report generation & CRIMSON $\uparrow$ & \textbf{0.8307} & 0.8204 \\
\hline
\multicolumn{4}{@{}l}{\textbf{(b) Image-resolution budget}} \\
Task & Metric & $768^2$ & $2048^2$ \\
\hline
Classification & BA $\uparrow$ & 0.7695 & \textbf{0.7791} \\
Multi-label classification & F1 $\uparrow$ & 0.6130 & \textbf{0.6286} \\
Detection & F1 $\uparrow$ & \textbf{0.9194} & 0.8914 \\
Regression & MAE $\downarrow$ & 10.1800 & \textbf{9.7800} \\
Counting & MAE $\downarrow$ & 13.4500 & \textbf{11.7300} \\
\hline
\end{tabular}
\end{table}

On Boneresorption, the specialist reduces MAE by 3.93 (paired 95\% confidence
interval $[0.83, 7.06]$) and produces 12 exact-zero answers versus 29 for the
general adapter. This supports the specialist recipe on that dataset, but does
not measure router accuracy.

\subsubsection{Image resolution.}

Raising the image budget from $768^2$ to $2048^2$ has mixed effects
(Table~\ref{tab:configuration_ablations}(b)). Classification BA increases by
0.0096, multi-label F1 increases by 0.0156, counting MAE decreases by 13\%,
and regression macro MAE decreases by 0.40, while detection F1 decreases by
0.0280. The paired 95\% interval for the
multi-label change is $[-0.004, +0.035]$, and both per-dataset regression
intervals include zero. Runs also used different GPUs and micro-batch shapes at
the same effective batch size, so the changes cannot be attributed to image
detail alone. Although the $768^2$ $A\cup B$ detection adapter scored 0.9194,
it is not the deployed route: the full-data detection adapter was trained at
$2048^2$. Table~\ref{tab:final_dataset_results} therefore reports the
route-matched $A\cup B$ proxy at $2048^2$ (0.8914 F1). Regression likewise
remains at $2048^2$, while report generation uses $768^2$.

\FloatBarrier

\subsection{Strengths and remaining failure modes}

An output-format audit of internal adapted-model predictions found valid nested
coordinate lists in all 588 detection responses, valid integers in all counting
responses, and no empty answers or leaked reasoning tags. The remaining errors
mainly reflect visual recognition and numerical estimation rather than syntax.

\begin{figure}[!htbp]
\centering
\begin{minipage}[t]{0.485\linewidth}
\centering
\includegraphics[width=\linewidth,height=3.4cm,keepaspectratio]{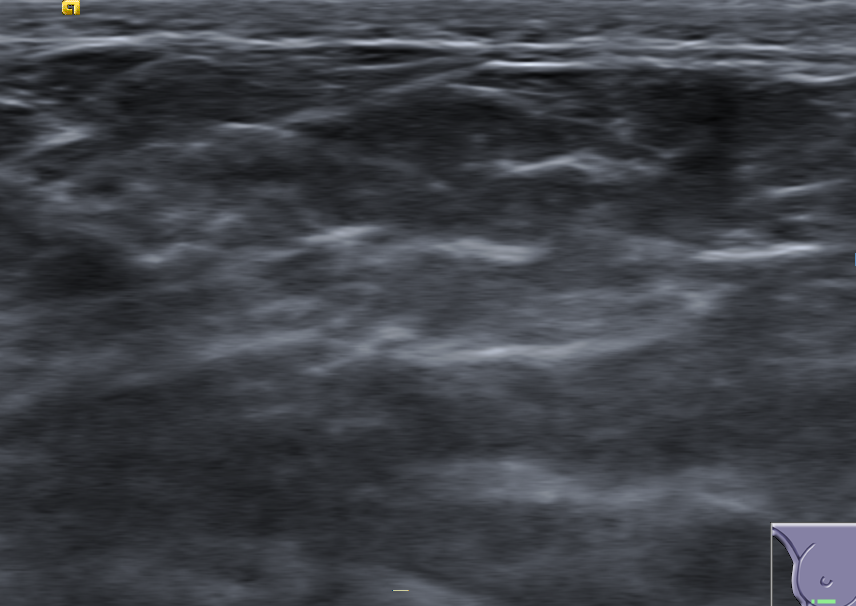}
\\[-0.8mm]\parbox[t]{\linewidth}{\raggedright\scriptsize
\textbf{(a) Classification, BUS-UCLM.}\\[-0.2mm]
\textbf{Question:} What does the contour-defined zone in the breast ultrasound scan signify? Options: A) Benign lesion present; B) Malignant lesion present; C) No lesion, only normal tissue. \textbf{Ground truth:} C. \textbf{Prediction:} C.}
\end{minipage}\hfill
\begin{minipage}[t]{0.485\linewidth}
\centering
\includegraphics[width=\linewidth,height=3.4cm,keepaspectratio]{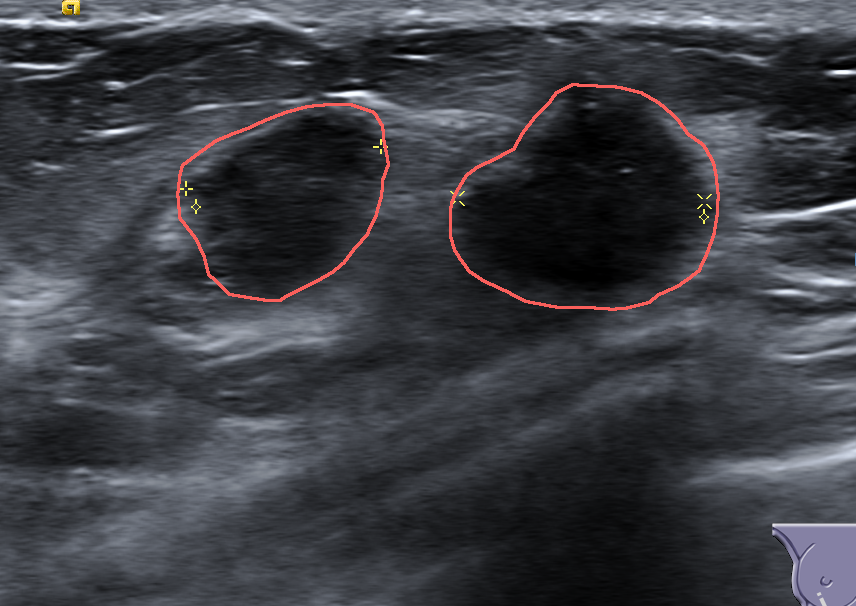}
\\[-0.8mm]\parbox[t]{\linewidth}{\raggedright\scriptsize
\textbf{(b) Classification, BUS-UCLM.}\\[-0.2mm]
\textbf{Question:} What does the contour-defined zone in the breast ultrasound scan signify? Options: A) Benign lesion present; B) Malignant lesion present; C) No lesion, only normal tissue. \textbf{Ground truth:} B. \textbf{Prediction:} A.}
\end{minipage}

\vspace{1mm}
\begin{minipage}[t]{0.485\linewidth}
\centering
\includegraphics[width=\linewidth,height=3.0cm,keepaspectratio]{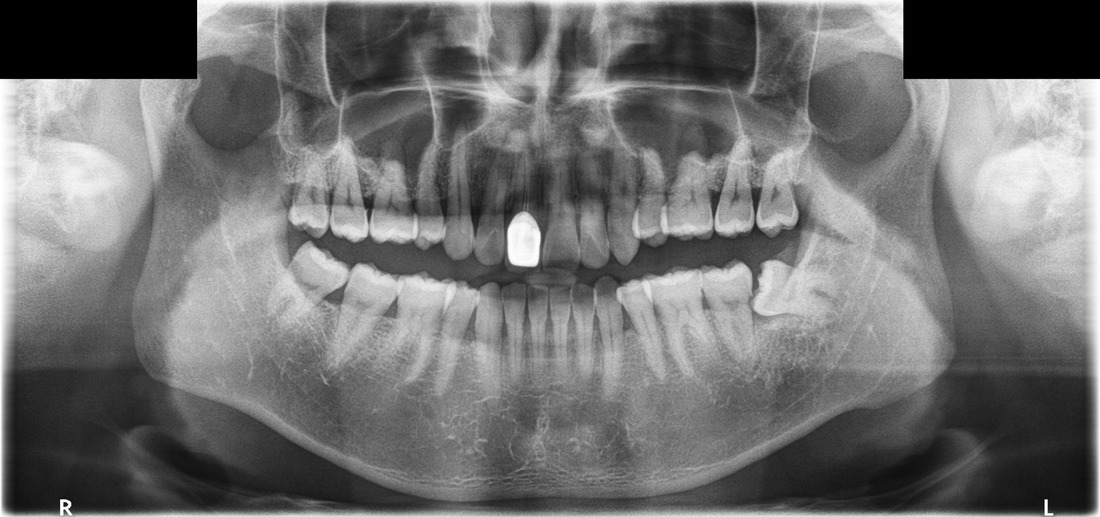}
\\[-0.8mm]\parbox[t]{\linewidth}{\raggedright\scriptsize
\textbf{(c) Regression, Boneresorption.}\\[-0.2mm]
\textbf{Question:} Assess the rate of alveolar bone resorption (ABR\%) from this orthopantomogram. \textbf{Ground truth:} 24.2. \textbf{Prediction:} 25.0. \textbf{Absolute error:} 0.8.}
\end{minipage}\hfill
\begin{minipage}[t]{0.485\linewidth}
\centering
\includegraphics[width=\linewidth,height=3.0cm,keepaspectratio]{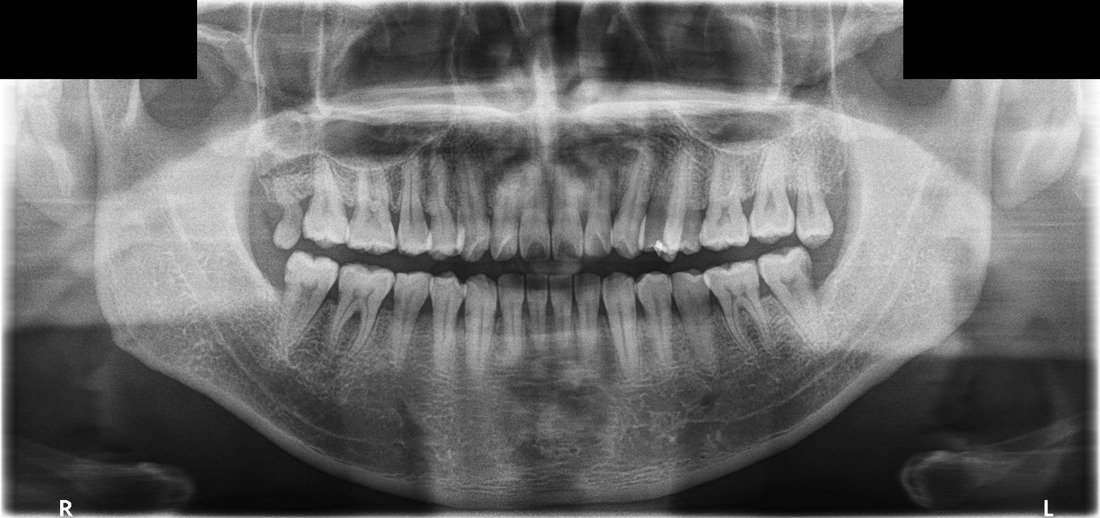}
\\[-0.8mm]\parbox[t]{\linewidth}{\raggedright\scriptsize
\textbf{(d) Regression, Boneresorption.}\\[-0.2mm]
\textbf{Question:} Estimate the alveolar bone resorption percentage (ABR\%) based on the radiographic findings in this panoramic image. \textbf{Ground truth:} 31.1. \textbf{Prediction:} 0.0. \textbf{Absolute error:} 31.1.}
\end{minipage}

\vspace{1mm}
\begin{minipage}[t]{0.485\linewidth}
\centering
\includegraphics[width=\linewidth,height=3.4cm,keepaspectratio]{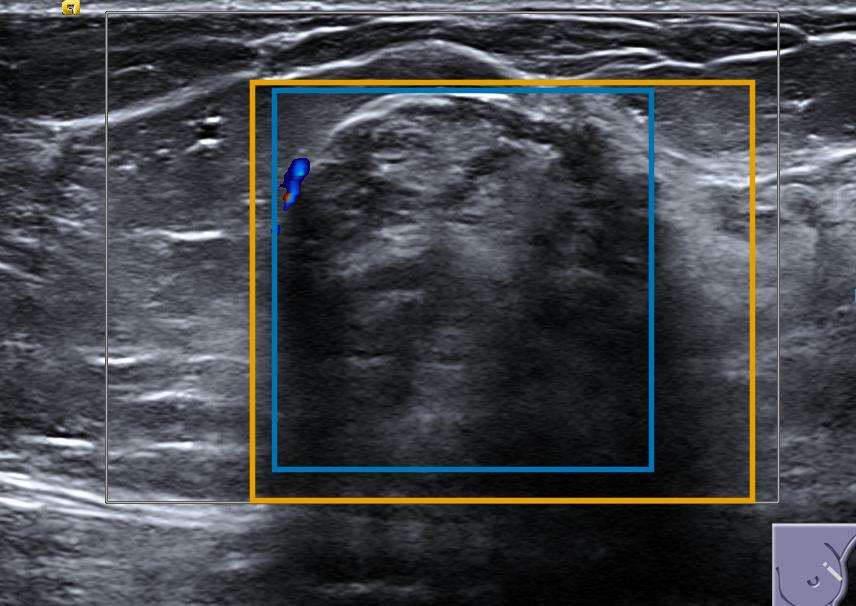}
\\[-0.8mm]\parbox[t]{\linewidth}{\raggedright\scriptsize
\textbf{(e) Detection, BUS-UCLM-det.} \textbf{IoU:} 0.68.\\[-0.2mm]
\textbf{Question:} Report the spatial coordinates of lesions in the ultrasound scan as $[x_{\min},y_{\min},x_{\max},y_{\max}]$, structuring multiple lesions as a nested list.\\[-0.2mm]
\textbf{Ground truth:} \texttt{[[274,90,651,469]]}.\\[-0.2mm]
\textbf{Prediction:} \texttt{[[252,82,752,500]]}.}
\end{minipage}\hfill
\begin{minipage}[t]{0.485\linewidth}
\centering
\includegraphics[width=\linewidth,height=3.4cm,keepaspectratio]{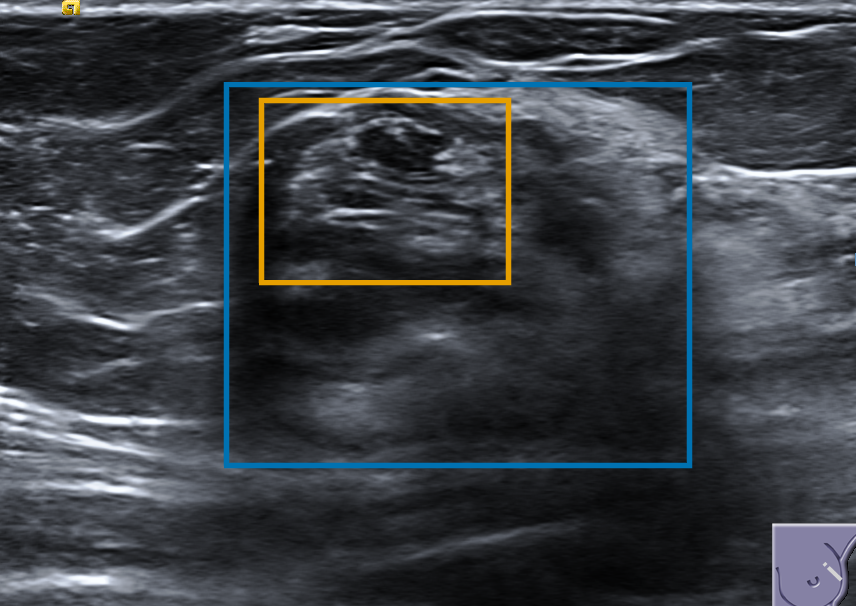}
\\[-0.8mm]\parbox[t]{\linewidth}{\raggedright\scriptsize
\textbf{(f) Detection, BUS-UCLM-det.} \textbf{IoU:} 0.25.\\[-0.2mm]
\textbf{Question:} Pinpoint lesion regions in the ultrasound image with $[x_{\min},y_{\min},x_{\max},y_{\max}]$ coordinates, nesting lists for multiple findings.\\[-0.2mm]
\textbf{Ground truth:} \texttt{[[226,84,689,465]]}.\\[-0.2mm]
\textbf{Prediction:} \texttt{[[261,100,508,282]]}.}
\end{minipage}
\vspace{-1mm}
\caption{Representative successes or low-error predictions (left) and failures
(right). Classification and detection use non-overlapping
\texttt{validation-public} cases; regression uses the internal holdout because
public validation has no regression cases. Blue boxes show ground truth and
orange boxes show predictions. All examples use task-specific $A\cup B$ adapters.}
\label{fig:qualitative_examples}
\end{figure}

Figure~\ref{fig:qualitative_examples} illustrates representative successes and
task-specific failures across categorical, numerical, and spatial outputs. All
displayed outputs are syntactically valid; the remaining errors arise from
visual category confusion, a degenerate numerical estimate, or incomplete
localization.

Report generation remains the main semantic limitation. Among 223 cases with
abnormal reference reports, 137 (61.4\%) receive normal-only predictions.
Only 16.4\% of predictions are unique, and the ten most frequent outputs
account for 60.6\%, indicating heavy reuse of report templates. The resulting
omissions are not apparent from the aggregate CRIMSON score of 0.8204.

\subsection{Deployment efficiency}
\label{sec:deployment}

The final container meets the runtime and size limits and produces identical outputs across two runs
(Table~\ref{tab:efficiency_results}).

\begin{table}[!t]
\centering
\caption{Final container verification on 32 cases covering all six tasks.}
\label{tab:efficiency_results}
\scriptsize
\setlength{\tabcolsep}{3pt}
\begin{tabular}{@{}p{4.8cm}ll@{}}
\hline
Measure & Observed & Limit or check \\
\hline
Mean runtime (incl. model loading) & 3.1 s/case & $\leq 6.57$ s/case \\
Container image size & 30.1 GB & $\leq 35$ GB \\
Repeated-run identity & 32/32 cases & Identical outputs \\
\hline
\end{tabular}
\end{table}

\subsection{Results on the Final Testing Set}

Table~\ref{tab:final_testing} reports the organizer-announced final-testing
results for our team, Hexapus, in the 2D track. These results are distinct
from the hidden-validation and internal dataset-macro results above.

\begin{table}[!htbp]
\centering
\caption{Organizer-reported final-testing results for Hexapus.}
\label{tab:final_testing}
\small
\setlength{\tabcolsep}{5pt}
\begin{tabular}{@{}rrrr@{}}
\hline
\shortstack{Balanced\\accuracy} & \shortstack{Multi-label\\F1} &
\shortstack{Detection\\F1} & MAE \\
\hline
0.8585 & 0.5310 & 0.8633 & 18.5286 \\
\hline
\end{tabular}
\end{table}

\subsection{Limitations and future work}

The organizer hidden-validation score covers classification, multi-label
classification, and regression only. The announced final-testing results additionally
reports detection F1, but no counting or report-generation scores. Evidence for
these two tasks is therefore limited to the internal and public-validation sets,
which differ in composition and are not directly comparable. Our Qwen3.5-9B
experiments also use a single training seed. The $2048^2$ production-adapter
runs peaked at approximately 40--44~GB of GPU memory, requiring a 48~GB-class or
larger GPU, whereas the lower-resolution runs fit on our 24~GB workstation GPUs;
the resolution comparisons also differed in GPU class and micro-batch
shape at the same effective batch size. Future work should confirm these
task-dependent effects across seeds, quantify errors introduced by the
question-only router, and investigate selective high-resolution processing or
adaptive tiling under matched compute budgets. A promising extension would combine our adapter routing with the multi-level visual feature
fusion of QwenVLConnector~\cite{nguyen2025qwenvlconnector}, letting the router also select a specialized visual pathway for fine spatial detail.

\section{Conclusion}

We demonstrate that a single quantized VLM can support six heterogeneous medical
image-parsing tasks without a separate backbone for each one. The submitted
30.1~GB system averaged 3.1 seconds per case and achieved an overall score of
0.4825 on organizer hidden validation across classification, multi-label
classification, and regression. Our controlled comparisons show no universal adaptation recipe: specialist adapters improved selected structured
tasks, higher resolution substantially benefited dense counting but produced
smaller or inconsistent gains elsewhere, and the general adapter remained
competitive for some routes. Report generation exposed a different limitation:
structurally valid outputs could still omit clinically relevant findings, with
61.4\% of abnormal-reference cases receiving normal-only predictions. A shared backbone should therefore unify deployment, not force uniform training
or inference; future generalist medical VLMs should allocate adapter capacity and
visual detail by task demand and be evaluated per dataset and failure mode, not aggregate scores.

\subsubsection{Acknowledgments} The authors of this paper declare that the proposed solution is fully automatic without any manual intervention. We acknowledge the Mohamed bin Zayed University of Artificial Intelligence Undergraduate Research Internship Program (MBZUAI UGRIP) for supporting this work. We thank all data owners and contributors for making the data publicly available and Codabench~\cite{codabench} for hosting the challenge platform.

\section*{Disclosure of Interests}
The authors declare no competing interests.

%
%
%
\bibliographystyle{splncs04}
\bibliography{paper}

\newpage
\begin{table}[!htbp]
\caption{Checklist Table. Please fill out this checklist table in the answer column.}
\centering
\begin{tabular}{ll}
\hline
Requirements & Answer \\ \hline

A meaningful title
& Yes \\ \hline

The number of authors ($\leq$6)
& 6 \\ \hline

Author affiliations and ORCID
& Yes \\ \hline

Corresponding author email is presented
& Yes \\ \hline

Validation scores are presented in the abstract
& Yes \\ \hline

\begin{tabular}[c]{@{}l@{}}
Introduction includes at least three parts: \\
background, related work, and motivation
\end{tabular}
& Yes \\ \hline

A pipeline/network figure is provided
& Figure~\ref{fig:pipeline} \\ \hline

Pre-processing
& Section~\ref{sec:data_preparation} \\ \hline

Strategies to use the partial label
& N/A \\ \hline

Strategies to use the unlabeled images.
& N/A \\ \hline

Strategies to improve model inference
& Section~\ref{sec:prompting} \\ \hline

Post-processing
& Section~\ref{sec:prompting} \\ \hline

The dataset and evaluation metric section are presented
& Section~\ref{sec:datasets} \\ \hline

Environment setting table is provided
& Table~\ref{table:env} \\ \hline

Training protocol table is provided
& Table~\ref{table:training} \\ \hline

Ablation study
& Table~\ref{tab:configuration_ablations} \\ \hline

Efficiency evaluation results are provided
& Table~\ref{tab:efficiency_results} \\ \hline

Visualized segmentation example is provided
& N/A (no segmentation task) \\ \hline

Limitation and future work are presented
& Yes \\ \hline

Reference format is consistent.
& Yes \\ \hline

\end{tabular}
\end{table}

\end{document}